\documentclass[11pt]{article}
\usepackage[preprint]{acl}
\usepackage[T1]{fontenc}
\usepackage[utf8]{inputenc}
\usepackage{times}
\usepackage{latexsym}
\usepackage{microtype}
\usepackage{graphicx}
\usepackage{booktabs}
\usepackage{amsmath}
\usepackage{amssymb}
\usepackage{multirow}
\usepackage{xcolor}

\newcommand{\NBItems}{348}
\newcommand{\NQItems}{1693}
\newcommand{\NBLns}{19}
\newcommand{\NQLns}{112}

\newcommand{\NBPhen}{70}
\newcommand{\NQPhen}{20}
\newcommand{\NQSame}{18}
\newcommand{\PctBLns}{5.5}
\newcommand{\PctQLns}{6.6}
\newcommand{\RelB}{0.537}
\newcommand{\RelQ}{0.454}
\newcommand{\NWeakB}{50}
\newcommand{\NWeakQ}{255}

\newcommand{\BPanelRecA}{0.47}

\newcommand{\BPanelAgrC}{0.934}
\newcommand{\BFcRecMin}{0.09}
\newcommand{\BFcRecMax}{0.20}
\newcommand{\BFcAgrMin}{0.937}
\newcommand{\BFcAgrMax}{0.953}
\newcommand{\BRateRecMin}{0.38}
\newcommand{\BRateRecMax}{0.52}
\newcommand{\BBest}{Gemini 3.8 Flash RATE}
\newcommand{\BBestF}{0.254}
\newcommand{\BBestDThree}{\ensuremath{-}0.077}
\newcommand{\BBestDThreeCI}{[\ensuremath{-}0.304, +0.150]}
\newcommand{\BNBelowOne}{6}

\newcommand{\QFcRecMin}{0.38}
\newcommand{\QFcRecMax}{0.46}
\newcommand{\QFcAgrMin}{0.903}
\newcommand{\QFcAgrMax}{0.917}
\newcommand{\QRateRecMin}{0.68}
\newcommand{\QRateRecMax}{0.75}
\newcommand{\QBest}{LLM panel (3$\times$FC)}
\newcommand{\QBestF}{0.366}

\newcommand{\QNBelowFiveSig}{5}
\newcommand{\NVal}{7}

\newcommand{\NWeakFcAll}{8}
\newcommand{\BPanelFlag}{1.7}
\newcommand{\FalseStrongRateMin}{5.4}
\newcommand{\FalseStrongRateMax}{23.2}

\newcommand{\NLm}{10}

\newcommand{\DeltaBar}{\ensuremath{-}18.9}
\newcommand{\DeltaCI}{[\ensuremath{-}22.1, \ensuremath{-}15.8]}
\newcommand{\DeltaBarAbs}{18.9}
\newcommand{\LiftDiff}{+5.6}
\newcommand{\LiftCI}{[+4.6, +6.7]}
\newcommand{\LiftLlm}{+5.6}
\newcommand{\LiftRand}{+0.0}
\newcommand{\NDeltaNeg}{10}
\newcommand{\DeltaOthMin}{\ensuremath{-}32.1}
\newcommand{\DeltaOthMax}{\ensuremath{-}18.2}

\newcommand{\HolmMax}{<0.001}
\newcommand{\DeltaStrong}{\ensuremath{-}16.4}
\newcommand{\DeltaStrongCI}{[\ensuremath{-}19.8, \ensuremath{-}13.2]}
\newcommand{\BDelta}{\ensuremath{-}15.7}
\newcommand{\BDeltaCI}{[\ensuremath{-}28.9, \ensuremath{-}2.7]}
\newcommand{\BNFlagSup}{23}
\newcommand{\BLmH}{+0.0}
\newcommand{\BLmHCI}{[\ensuremath{-}1.3, +1.4]}

\newcommand{\BRemovedLlm}{31}

\newcommand{\QDelta}{\ensuremath{-}19.1}
\newcommand{\QDeltaCI}{[\ensuremath{-}22.4, \ensuremath{-}15.9]}
\newcommand{\QNFlagSup}{341}
\newcommand{\QLmH}{+3.4}
\newcommand{\QLmHCI}{[+2.6, +4.3]}

\newcommand{\QRemovedLlm}{426}

\newcommand{\QRemovedBoth}{85}
\newcommand{\QLnsPassed}{27}

\newcommand{\NComboNeg}{70}

\newcommand{\DeltaLen}{\ensuremath{-}14.5}

\newcommand{\LooMin}{\ensuremath{-}20.3}
\newcommand{\LooMax}{\ensuremath{-}17.5}
\newcommand{\TauMin}{0.87}
\newcommand{\FlipMax}{2}

\newcommand{\DsQNone}{0.491}
\newcommand{\CtlHOne}{\ensuremath{-}13.7}

\newcommand{\CtlHThree}{\ensuremath{-}21.7}

\newcommand{\CtlHOneRate}{11.4}
\newcommand{\CtlHThreeRate}{5.9}
\newcommand{\CtlVsOneMin}{\ensuremath{-}5.2}
\newcommand{\CtlVsOneMax}{\ensuremath{-}18.5}

\newcommand{\CtlFcVsThreeMin}{\ensuremath{-}5.8}
\newcommand{\CtlFcVsThreeMax}{\ensuremath{-}10.5}
\newcommand{\CtlFcVsThreeSig}{3}
\newcommand{\CtlNFc}{4}
\newcommand{\CtlFcRateMin}{4.3}
\newcommand{\CtlFcRateMax}{6.2}

\newcommand{\CtlRateVsThreeMin}{\ensuremath{-}0.5}
\newcommand{\CtlRateVsThreeMax}{+2.8}
\newcommand{\CtlRateRateMin}{19.4}
\newcommand{\CtlRateRateMax}{24.6}

\newcommand{\VerBRecMin}{0.28}
\newcommand{\VerBRecMax}{0.40}
\newcommand{\VerBFMin}{0.153}
\newcommand{\VerBFMax}{0.244}

\newcommand{\VerQFMin}{0.229}
\newcommand{\VerQFMax}{0.321}

\newcommand{\VerDeltaMin}{\ensuremath{-}14.2}
\newcommand{\VerDeltaMax}{\ensuremath{-}18.9}

\newcommand{\VerLmAll}{30}
\newcommand{\VerLiftMin}{+2.7}
\newcommand{\VerLiftMax}{+4.5}
\newcommand{\VerVsOneMin}{\ensuremath{-}0.9}
\newcommand{\VerVsOneMax}{\ensuremath{-}5.7}

\newcommand{\VerVsThreeLo}{+2.2}
\newcommand{\VerVsThreeHi}{+7.1}

\newcommand{\NNlItems}{756}
\newcommand{\NNlPhen}{84}
\newcommand{\NNlLns}{38}

\newcommand{\NlRatGood}{20}
\newcommand{\NlRatBad}{10}

\newcommand{\NlDelta}{\ensuremath{-}13.4}
\newcommand{\NlDeltaAbs}{13.4}
\newcommand{\NlDeltaCI}{[\ensuremath{-}21.1, \ensuremath{-}5.4]}
\newcommand{\NlLmNeg}{10}
\newcommand{\NlFlagSup}{50}
\newcommand{\NlLift}{+1.6}
\newcommand{\NlLiftCI}{[+0.7, +2.5]}
\newcommand{\NlOthMin}{\ensuremath{-}47.0}
\newcommand{\NlOthMax}{\ensuremath{-}8.7}
\newcommand{\NlHolmSig}{5}
\newcommand{\NlNOth}{6}
\newcommand{\NlComboNeg}{66}
\newcommand{\NlCombo}{70}
\newcommand{\NlStrong}{\ensuremath{-}14.8}
\newcommand{\NlStrongCI}{[\ensuremath{-}30.0, +0.8]}
\newcommand{\NlLen}{\ensuremath{-}16.5}
\newcommand{\NlLenCI}{[\ensuremath{-}24.1, \ensuremath{-}9.1]}
\newcommand{\NlLooMin}{\ensuremath{-}15.2}
\newcommand{\NlLooMax}{\ensuremath{-}11.3}
\newcommand{\NlLmH}{+0.1}
\newcommand{\NlLmHCI}{[\ensuremath{-}0.7, +0.9]}

\newcommand{\NlPanelRecA}{0.49}

\newcommand{\NlPanelFB}{0.430}

\newcommand{\NlFcRecMin}{0.15}
\newcommand{\NlFcRecMax}{0.43}
\newcommand{\NlBest}{Gemini 3.8 Flash RATE}
\newcommand{\NlBestF}{0.388}

\newcommand{\NlCtlHOne}{\ensuremath{-}8.5}
\newcommand{\NlCtlHOneCI}{[\ensuremath{-}14.7, \ensuremath{-}2.7]}
\newcommand{\NlCtlVsOneNeg}{7}
\newcommand{\NlCtlVsOneSig}{1}

\title{LLM Validation Makes Minimal-Pair Benchmarks Easier for Language Models}

\author{Guofang Tang \\
  \texttt{tinggofun@gmail.com} \And
  Zheng Qu \\
  Fudan University \\
  \texttt{1262333286@qq.com}}

\begin{document}
\maketitle

\begin{abstract}
Minimal-pair benchmarks are only as good as their labels, and LLMs are an obvious substitute for costly human validation.
We test it on BLiMP (English) and QFrBLiMP (Quebec French), the two public minimal-pair datasets releasing every forced-choice human judgment.
Three LLMs validate by forced choice, by rating, and as a panel, scored against human panels of one, three and five judges in two pre-registered phases.
On BLiMP, forced-choice validators agree with held-out human majorities more often than five-judge panels, yet catch far fewer pairs whose label humans reject than one judge does.
By point estimate, no validator is worth more than one judge.
The errors are not random.
On both datasets, across \NLm{} open LMs, human-supported pairs that the pre-registered LLM filter flags are \DeltaBarAbs{} points harder than those it accepts, in all \NComboNeg{} LM--validator combinations, and harder than pairs a single human judge flags.
Filtering thus removes items on which LMs fail but humans do not, raising accuracy more than random removal.
A protocol frozen before any output replicates the gap and the lift on Dutch BLiMP-NL.
LLM validation makes minimal-pair benchmarks easier for the models they are meant to test; we release code, validator outputs and item-level scores.
\end{abstract}

\section{Introduction}
\label{sec:intro}

A minimal pair puts two sentences side by side that differ in one grammatical respect. One is meant to be acceptable
and the other not. If a language model gives the acceptable one a higher probability, it gets credit for the
phenomenon \citep{warstadt2020blimp}. The method has spread from English to Russian, Quebec French and more than
a hundred languages \citep{taktasheva2024rublimp,beauchemin2026qfrblimp,jumelet2025multiblimp}. Its
weak point is the label. Templates and generators produce pairs whose intended contrast native speakers do not share,
and acceptability is graded, so speakers disagree \citep{sprouse2013comparison,lau2017grammaticality}. Benchmark
authors therefore validate labels with human judges. That step is slow, costs money, and is hard to staff in
low-resource languages.

LLMs look like a cheap substitute. They annotate text about as well as crowd workers on many tasks
\citep{gilardi2023chatgpt}, they serve as judges for generated text \citep{zheng2023judging}, and on English
grammatical constructions they sometimes match the intended label more often than humans do \citep{hu2024language}.
There is now a test for whether an LLM can stand in for a human annotator \citep{calderon2025alternative}. For
minimal pairs, though, the substitution has a specific risk. The models that would validate the benchmark belong to
the same family as the models the benchmark is meant to test. If both prefer the same sentences, LLM validation
selects items on the outcome it is supposed to measure.

We test this on two minimal-pair datasets that release every individual human judgment: the
\NBItems{} pairs of BLiMP that were rated by about 20 crowd workers each, and all \NQItems{} pairs of QFrBLiMP, each
judged by 12 native speakers of Quebec French. A pair's \emph{support} is the share of judges who choose the intended
acceptable sentence. We call a pair LNS (label not supported) when at most half of the judges choose it. Three LLMs
from different developers validate every pair, by forced choice and by rating each sentence on its own, and a
three-model panel votes. In two phases, each locked before any of its outputs existed, we ask two questions.

\begin{figure*}[t]
\centering
\includegraphics[width=\textwidth]{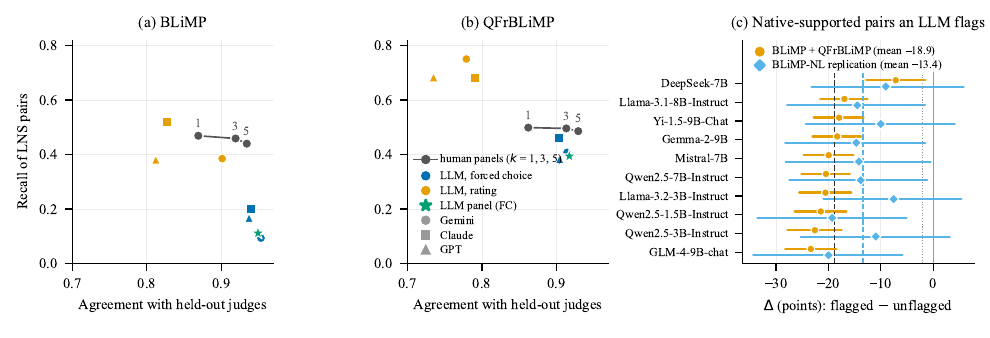}
\caption{(a, b) LLM validators against human panels. The y-axis shows how many of the pairs whose intended label the
held-out judges do not support (LNS) a validator flags. The x-axis shows how often it agrees with the held-out
majority on which sentence is better. Forced-choice validators (blue) and the LLM panel (green) sit at high
agreement and low recall, below a single human (``1''). Rating validators (orange) recall more but flag many
supported pairs. (c) For each open LM: accuracy on human-supported pairs that the pre-registered LLM filter (Gemini
RATE) flags, minus accuracy on those it accepts, in Phase~1 (orange) and in the frozen BLiMP-NL replication (blue).
Bars show 95\% item-bootstrap CIs, dashed lines the means and the dotted line the pre-registered $-2$-point
threshold.}
\label{fig:main}
\end{figure*}

\paragraph{Can an LLM validator replace human judges?} We compare each validator with human panels of one, three and
five judges, all scored against the remaining judges. On BLiMP, forced-choice validators agree with the held-out
majority more often than five-judge panels do, yet they detect few LNS pairs (Figure~\ref{fig:main}a,b). By point estimate, no validator matches
more than one human judge. The pre-registered test on BLiMP, which has only \NBLns{} LNS pairs, was a weak pass: the
best validator fell below the three-judge panel, but the interval included zero.

\paragraph{Does LLM validation change what we conclude about LMs?} We score both datasets with \NLm{} open LMs from
1.5B to 9B parameters. Among pairs that humans support, the pairs an LLM validator flags are much harder for LMs than
the pairs it accepts (Figure~\ref{fig:main}c), and on these two datasets harder than the pairs a single human judge
flags. Removing the flagged
pairs raises LM accuracy beyond what removing random pairs does. Both pre-registered conditions hold. They hold
again in a third protocol on Dutch BLiMP-NL \citep{suijkerbuijk2025blimpnl}, with every prompt, filter and criterion
frozen before any output existed.

\medskip\noindent Our contributions are:
\begin{itemize}\setlength\itemsep{0pt}
\item A human-panel comparison for LLM validation of minimal pairs, scored against held-out judges. LLM validators
show \emph{high agreement and low detection}: forced-choice validators agree with the held-out majority on
\BFcAgrMin--\BFcAgrMax{} of BLiMP pairs, above the five-judge panel (\BPanelAgrC), but recover only
\BFcRecMin--\BFcRecMax{} of the LNS pairs, against \BPanelRecA{} for one judge.
\item Evidence that LLM validation is biased in the models' favour. Human-supported pairs that the pre-registered
LLM filter flags are \DeltaBarAbs{} points harder for open LMs ($\bar\Delta = \DeltaBar$, 95\% CI \DeltaCI). The
gap is negative for all \NComboNeg{} combinations of LM and validator, and larger than for pairs that a single human
judge flags. Filtering with the LLM raises accuracy by
\LiftDiff{} points more than random filtering. On QFrBLiMP, every LM scores higher on the LLM-filtered benchmark than
on the human-filtered one. A frozen replication on Dutch reproduces the gap ($\bar\Delta = \NlDelta$, \NlDeltaCI)
and the lift.
\item Three pre-registered protocols, the code, and all validator outputs.
\end{itemize}

\section{Data, gold labels and validators}
\label{sec:setup}

\paragraph{Datasets.} We use the two public minimal-pair datasets we found that release every individual forced-choice
judgment. A third, with graded ratings, is used for the replication in Section~\ref{sec:replication}.
\begin{itemize}\setlength\itemsep{0pt}
\item \textbf{BLiMP} \citep{warstadt2020blimp}, English. The authors had \NBItems{} pairs, about five per paradigm (\NBPhen{} paradigm
labels in the released validation file), judged on Amazon Mechanical Turk by 19--21 raters per pair. Raters chose the more natural sentence of each pair.
\item \textbf{QFrBLiMP} \citep{beauchemin2026qfrblimp}, Quebec French. Twelve native speakers judged all 1,711 pairs
from \NQPhen{} phenomena, choosing the sentence they found well written. We drop \NQSame{} pairs whose two sentences
are identical in the release, which leaves \NQItems{}.\footnote{In the released files the \texttt{grammatical} and
\texttt{ungrammatical} fields are swapped; the authors' \texttt{generate\_truth.py} notes this. With the swap undone,
mean support is 0.86, as reported by the authors.}
\end{itemize}

\paragraph{Gold labels.} For each pair, the support $s$ is the share of judges who chose the intended acceptable
sentence. A pair is \textbf{LNS} (label not supported) if $s \le 0.5$. This gives \NBLns{} LNS pairs in BLiMP
(\PctBLns\%) and \NQLns{} in QFrBLiMP (\PctQLns\%). LNS does not mean that a pair is invalid: the judges may reflect
dialect variation, graded acceptability or a task effect. It means only that the benchmark's label is not what the
judges chose. A secondary gold uses weak support, $s<0.75$ (\NWeakB{} and \NWeakQ{} pairs). Split-half reliability of
$s$, as a Spearman correlation between random halves of the judges, is \RelB{} for BLiMP and \RelQ{} for QFrBLiMP.

\paragraph{Human panels and held-out gold.} To ask how many humans an LLM validator is worth, we build human panels
from the same data. In each of 1,000 resamples, we shuffle each pair's judges. The first one, three and five judges
form panels of size $k=1,3,5$, and a panel flags a pair if its majority does not choose the intended sentence. The
remaining judges (BLiMP 14--16, QFrBLiMP 7) give a held-out support $s'$ and a held-out gold. The three panels and
every LLM validator are scored against the same held-out gold in each resample. We report LNS recall, precision and
F1, and \emph{agreement}: the share of pairs on which a validator prefers the same sentence as the held-out
majority. A validator's \textbf{equivalent panel size} $k_{\text{eq}}$ is the largest $k$ whose panel F1 is not above the
validator's, by point estimate; ``${<}1$'' means below a single judge.
Intervals for differences come from 1,000 paired item bootstraps, which redraw the items and the judge order
together.

\paragraph{LLM validators.} We use three LLMs from different developers through their APIs, at temperature 0:
Gemini 3.8 Flash, Claude Sonnet 4.6 and GPT-6 Luna. Each validates every pair in two ways (prompts in
Appendix~\ref{app:prompts}).
\begin{itemize}\setlength\itemsep{0pt}
\item \textbf{FC} (forced choice) reuses the human task's instruction. Twelve pairs go into each request, in a fixed
random order of sentences. A validator flags a pair if it chooses the unintended sentence. The French prompt asks
the model to answer as a native speaker of Quebec French.
\item \textbf{RATE} asks for a 1--5 acceptability rating of each sentence on its own. Sentences are shuffled into
batches of 12, and the two sentences of a pair never share a batch. A validator flags a pair if the intended
sentence is not rated higher ($r_{\text{good}} \le r_{\text{bad}}$).
\end{itemize}
The \textbf{LLM panel} is the majority of the three FC validators, giving seven validators in all. Every request
returned a parseable answer.\footnote{One batch of 12 BLiMP sentences repeatedly returned empty content from one
model; it was split in halves and re-sent.}

\paragraph{Pre-registration.} Phase~0 (this section and Section~\ref{sec:validators}) was locked before any
validator output existed, and Phase~1 (Section~\ref{sec:downstream}) before any LM was scored. Each protocol fixes
the gold, the estimands, the decision rules and the analyses that are only exploratory. A clarification of the
Phase~0 resampling, added before any output was read, is listed with the protocols in the supplement.

\section{Can an LLM validator replace human judges?}
\label{sec:validators}

\begin{table*}[t]
\centering
\small
\resizebox{\textwidth}{!}{%
\begin{tabular}{l ccccc ccccc}
\toprule
 & \multicolumn{5}{c}{BLiMP (English, 348 pairs, 19 LNS)} & \multicolumn{5}{c}{QFrBLiMP (Quebec French, 1693 pairs, 112 LNS)} \\
\cmidrule(lr){2-6}\cmidrule(lr){7-11}
Validator & Rec & Prec & F1 & Agree & $k_{\text{eq}}$ & Rec & Prec & F1 & Agree & $k_{\text{eq}}$ \\
\midrule
1 judge & 0.47 & 0.16 & 0.239 & 0.869 & -- & 0.50 & 0.22 & 0.304 & 0.862 & -- \\
3 judges & 0.46 & 0.26 & 0.331 & 0.919 & -- & 0.50 & 0.35 & 0.409 & 0.913 & -- \\
5 judges & 0.44 & 0.32 & 0.365 & 0.934 & -- & 0.49 & 0.43 & 0.454 & 0.929 & -- \\
\midrule
Gemini 3.8 Flash FC & 0.09 & 0.36 & 0.148 & 0.953 & $<$1 & 0.41 & 0.33 & 0.364 & 0.914 & 1 \\
Gemini 3.8 Flash RATE & 0.39 & 0.19 & 0.254 & 0.901 & 1 & 0.75 & 0.18 & 0.291 & 0.779 & $<$1 \\
Claude Sonnet 4.6 FC & 0.20 & 0.26 & 0.223 & 0.940 & $<$1 & 0.46 & 0.30 & 0.364 & 0.903 & 1 \\
Claude Sonnet 4.6 RATE & 0.52 & 0.13 & 0.208 & 0.827 & $<$1 & 0.68 & 0.18 & 0.281 & 0.790 & $<$1 \\
GPT-6 Luna FC & 0.17 & 0.21 & 0.186 & 0.937 & $<$1 & 0.38 & 0.28 & 0.325 & 0.904 & 1 \\
GPT-6 Luna RATE & 0.38 & 0.09 & 0.150 & 0.812 & $<$1 & 0.68 & 0.14 & 0.238 & 0.735 & $<$1 \\
LLM panel (3$\times$FC) & 0.11 & 0.29 & 0.160 & 0.949 & $<$1 & 0.39 & 0.34 & 0.366 & 0.917 & 1 \\
\bottomrule
\end{tabular}}
\caption{Finding the pairs whose intended label natives do not support (LNS). Every validator is scored against the held-out judges of the same resample (1{,}000 judge reshuffles). Rec/Prec/F1: LNS detection. Agree: agreement with the held-out majority on which sentence is better. $k_{\text{eq}}$: the largest human panel whose F1 the validator matches. FC: forced choice; RATE: 1--5 rating of each sentence on its own. QFrBLiMP is reported but not used for the pre-registered verdict, because its split-half reliability is below 0.5 (0.454).}
\label{tab:phase0}
\end{table*}

Table~\ref{tab:phase0} and Figure~\ref{fig:main}a,b give the results. Three patterns are visible in both datasets.
The pre-registered verdict uses BLiMP only (end of this section); QFrBLiMP results are descriptive.

\paragraph{High agreement, low detection.} Forced-choice validators almost always side with the benchmark's intended
label. On BLiMP their agreement with the held-out majority is \BFcAgrMin--\BFcAgrMax, higher than that of a
five-judge panel (\BPanelAgrC). Yet they recover only \BFcRecMin--\BFcRecMax{} of the LNS pairs, where a single judge
recovers \BPanelRecA. The LLM panel flags \BPanelFlag\% of BLiMP pairs, although \PctBLns\% are LNS. High agreement is
what one would see from a validator that rarely disagrees with the benchmark authors. Over a benchmark where most labels
are right, that looks accurate, and it misses the pairs that matter. On QFrBLiMP the forced-choice validators recall
more (\QFcRecMin--\QFcRecMax), but their agreement (\QFcAgrMin--\QFcAgrMax) still falls short of the five-judge
panel.

\paragraph{Rating trades misses for false alarms.} Rating each sentence on its own recalls more LNS pairs (BLiMP
\BRateRecMin--\BRateRecMax, QFrBLiMP \QRateRecMin--\QRateRecMax), but precision drops. Rating validators flag
\FalseStrongRateMin--\FalseStrongRateMax\% of the pairs that at least three quarters of the judges support. A benchmark
author who deletes every flagged pair would lose that share of good items. Section~\ref{sec:downstream} shows that the
lost items are not a random share.

\paragraph{No validator is worth more than one judge.} On BLiMP, \BNBelowOne{} of the \NVal{} validators fall below a
single judge in F1. The best, \BBest{} (F1 \BBestF), reaches one judge but not three. On QFrBLiMP, the best validator
(\QBest, F1 \QBestF) also reaches one judge but not three, and \QNBelowFiveSig{} of the \NVal{} validators are below the five-judge
panel with intervals that exclude zero. With the weak-support gold ($s<0.75$, Appendix~\ref{app:secondary}), all
\NWeakFcAll{} comparisons of a forced-choice validator with the three-judge panel favour the panel, with intervals
excluding zero.

\paragraph{Showing the label does not help.} A benchmark builder might instead show the intended label and ask whether
native speakers would agree. We added this VERIFY prompt after Phase~1 (Appendix~\ref{app:verify}). On BLiMP it
recovers more LNS pairs than forced choice (\VerBRecMin--\VerBRecMax) but loses precision. Its F1 (BLiMP
\VerBFMin--\VerBFMax, QFrBLiMP \VerQFMin--\VerQFMax) is never above that of a single judge.

\paragraph{The pre-registered verdict.} The Phase~0 rule judged BLiMP only, because QFrBLiMP's split-half reliability
(\RelQ) is below the pre-set threshold of 0.5. It required the best validator to fall below the three-judge panel
with an interval excluding zero. The best validator's F1 minus the panel's is \BBestDThree, but with \NBLns{} LNS
pairs the interval is wide (\BBestDThreeCI). The kill condition (the best validator matching the five-judge panel) is
not met either, so the verdict is the pre-defined \emph{weak pass}. The Phase~1 test does not depend on this
estimate; it asks whether the validators' errors are systematic.

\section{Does LLM validation change what we conclude about LMs?}
\label{sec:downstream}

A validator that misses LNS pairs and flags supported ones is a problem only if its errors are systematic. The risk
specific to minimal pairs is that the validator's errors line up with the errors of the LMs being evaluated.

\paragraph{Setup.} We score all pairs with \NLm{} open LMs from seven developers: Qwen2.5-1.5B/3B/7B-Instruct,
Llama-3.2-3B-Instruct, Llama-3.1-8B-Instruct, GLM-4-9B-chat, Mistral-7B, Yi-1.5-9B-Chat, DeepSeek-7B and Gemma-2-9B.
Following \citet{warstadt2020blimp}, an LM gets a pair right if the intended sentence has the higher total
log-probability. The pre-registered primary filter is Gemini RATE, the validator with the highest F1 on BLiMP in
Phase~0. Only pairs that humans support ($s>0.5$) enter the primary estimand. Within each dataset and for each LM,
\[
\Delta = \text{acc}(\text{flagged}) - \text{acc}(\text{not flagged}).
\]
We combine the two datasets with weights fixed by their numbers of flagged supported pairs (\BNFlagSup{} and
\QNFlagSup), and average over LMs to get $\bar\Delta$. The datasets are not pooled directly, because flag rates and LM
accuracy both differ by language. The second estimand is the \emph{lift}: accuracy after removing all flagged pairs
minus accuracy on the full set, compared with removing the same number of random pairs. Intervals come from 10,000
item bootstraps, stratified by dataset and shared across LMs. Accuracy differences are reported in percentage
points. The claim was to count as supported only if
$\bar\Delta \le -2$ points with a 95\% interval below zero \emph{and} the lift exceeded random removal with an interval
above zero.

\begin{table*}[t]
\centering
\small
\begin{tabular}{l c r c c r c}
\toprule
 & \multicolumn{3}{c}{Phase 1: BLiMP + QFrBLiMP} & \multicolumn{3}{c}{Replication: BLiMP-NL} \\
\cmidrule(lr){2-4}\cmidrule(lr){5-7}
Filter & Flagged & $\bar\Delta$ & 95\% CI & Flagged & $\bar\Delta$ & 95\% CI \\
\midrule
Gemini FC & 2 / 80 & \ensuremath{-}30.9 & [\ensuremath{-}37.4, \ensuremath{-}24.5] & 6 & \ensuremath{-}47.0 & [\ensuremath{-}64.8, \ensuremath{-}25.0] \\
Gemini RATE$^\dagger$ & 23 / 341 & \ensuremath{-}18.9 & [\ensuremath{-}22.1, \ensuremath{-}15.8] & 50 & \ensuremath{-}13.4 & [\ensuremath{-}21.1, \ensuremath{-}5.4] \\
Claude FC & 9 / 104 & \ensuremath{-}29.4 & [\ensuremath{-}35.3, \ensuremath{-}23.6] & 41 & \ensuremath{-}16.7 & [\ensuremath{-}25.4, \ensuremath{-}8.0] \\
Claude RATE & 51 / 315 & \ensuremath{-}20.6 & [\ensuremath{-}23.7, \ensuremath{-}17.5] & 114 & \ensuremath{-}11.4 & [\ensuremath{-}17.1, \ensuremath{-}6.0] \\
GPT-6 FC & 9 / 93 & \ensuremath{-}27.1 & [\ensuremath{-}33.4, \ensuremath{-}21.0] & 29 & \ensuremath{-}11.0 & [\ensuremath{-}20.1, \ensuremath{-}2.1] \\
GPT-6 RATE & 55 / 407 & \ensuremath{-}18.2 & [\ensuremath{-}21.1, \ensuremath{-}15.4] & 102 & \ensuremath{-}8.7 & [\ensuremath{-}14.3, \ensuremath{-}3.1] \\
LLM panel & 4 / 71 & \ensuremath{-}32.1 & [\ensuremath{-}39.7, \ensuremath{-}24.7] & 7 & \ensuremath{-}19.2 & [\ensuremath{-}40.9, +1.4] \\
\bottomrule
\end{tabular}
\caption{How much harder the pairs that a validator flags, but natives support, are for open LMs. $\bar\Delta$: LM accuracy on flagged minus unflagged native-supported pairs, in points, averaged over 10 LMs. Flagged: native-supported pairs flagged (Phase 1: BLiMP / QFrBLiMP). Gemini 3.8 Flash, Claude Sonnet 4.6, GPT-6 Luna; LLM panel = majority of the three FC validators. $^\dagger$Pre-registered primary filter; for the other six, Holm-adjusted $p<0.001$ in Phase 1. BLiMP-NL: same validators, LMs and estimand, all locked before any output; support from graded ratings (Section~\ref{sec:replication}).}
\label{tab:downstream}
\end{table*}

\begin{table*}[t]
\centering
\small
\begin{tabular}{l l p{7.5cm} c c c}
\toprule
Type & Data & \emph{Intended sentence} / other sentence & $s$ & Flagged & LM acc. \\
\midrule
Supported, flagged & QFr & \emph{Ce peintre emploie judicieusement les matériaux que sont l'huile, l'acrylique et l'aquarelle.} / Ce peintre emploie judicieusement les médiums que sont l'huile, l'acrylique et l'aquarelle. & 0.92 & 7/7 & 0.2 \\
Supported, flagged & QFr & \emph{Les copinages ont occupé la moitié de l'émission!} / Les plugs ont occupé la moitié de l'émission! & 0.92 & 6/7 & 0.2 \\
LNS, kept & BLiMP & \emph{James is pleasant to flee from.} / James is apt to flee from. & 0.50 & 0/7 & 0.5 \\
LNS, kept & QFr & \emph{La majorité des amis de Pierre organisent une fête pour souligner son anniversaire.} / La majorité des amis de Pierre organise une fête pour souligner son anniversaire. & 0.50 & 0/7 & 0.5 \\
\bottomrule
\end{tabular}
\caption{Example pairs. $s$: share of judges choosing the intended sentence. Flagged: how many of the seven validators flag the pair. LM acc.: share of the ten LMs that prefer the intended sentence. In the first two rows, the native judges prefer the intended word, while the LLM validators do not and most LMs prefer the other word. In the last two rows, half of the judges reject the intended label, and no validator notices.}
\label{tab:examples}
\end{table*}

\paragraph{Flagged pairs are much harder for LMs.} Both conditions hold. Human-supported pairs that the primary filter
flags are \DeltaBarAbs{} points harder than those it accepts ($\bar\Delta = \DeltaBar$, \DeltaCI), and $\Delta<0$ for
all \NDeltaNeg{} LMs (Figure~\ref{fig:main}c). The other six validators show the same gap, from \DeltaOthMax{} to
\DeltaOthMin{} points (Table~\ref{tab:downstream}; Holm-adjusted $p\HolmMax$). Forced-choice validators flag fewer
pairs, but the pairs they flag are even harder. $\Delta$ is negative for all \NComboNeg{} combinations of LM and
validator. The gap holds on each dataset (BLiMP \BDelta, \BDeltaCI; QFrBLiMP \QDelta, \QDeltaCI). It also holds when
only strongly supported pairs ($s \ge 0.75$) count (\DeltaStrong, \DeltaStrongCI), with length-normalised
log-probabilities (\DeltaLen), and when any single phenomenon is left out (\LooMin{} to \LooMax). The post-hoc
VERIFY prompt gives the same picture: \VerDeltaMin{} to \VerDeltaMax{} points, every interval below zero, negative
for all \VerLmAll{} combinations of LM and validator, and a lift over random removal of \VerLiftMin{} to
\VerLiftMax{} points (Appendix~\ref{app:verify}).

\paragraph{Not just hard pairs.} Perhaps any judge's disagreement marks pairs that are hard for LMs. A post-hoc
control tests this (Table~\ref{tab:control}). In each judge reshuffle, the first judge, or the majority of the first
three, flags pairs, and support comes from the held-out judges. Human flags do mark hard pairs: \CtlHOne{} points for
one judge and \CtlHThree{} for three. LLM flags mark harder ones. For all \NVal{} validators the gap is larger
than for one judge (differences \CtlVsOneMin{} to \CtlVsOneMax{} points, every interval below zero), even though
rating validators flag about twice as many pairs (\CtlRateRateMin--\CtlRateRateMax\% against \CtlHOneRate\%), which
would normally dilute the gap. The forced-choice validators and the LLM panel flag about as many pairs as three judges
(\CtlFcRateMin--\CtlFcRateMax\% against \CtlHThreeRate\%), and their gap is larger by \CtlFcVsThreeMin{} to
\CtlFcVsThreeMax{} points (\CtlFcVsThreeSig{} of \CtlNFc{} intervals below zero). Rating validators match the
three-judge panel (\CtlRateVsThreeMin{} to \CtlRateVsThreeMax). Under the two pre-registered prompts, LLM errors are
thus more aligned with LM errors than the disagreement of one judge, and for forced choice more than that of three.
VERIFY flags more pairs, and its gap is close to that of one judge (differences \VerVsOneMin{} to \VerVsOneMax,
intervals including zero) and smaller than that of three (\VerVsThreeLo{} to \VerVsThreeHi).

\paragraph{LLM filtering inflates scores.} Removing the pairs the primary filter flags raises LM accuracy by
\LiftLlm{} points, while removing the same number of random pairs changes it by \LiftRand{} (difference \LiftDiff,
\LiftCI). Human filtering raises scores too, since LNS pairs are hard for LMs as well. But the two filters remove
different pairs. On QFrBLiMP the LLM filter removes \QRemovedLlm{} pairs, of which only \QRemovedBoth{} are LNS, and
it keeps \QLnsPassed{} of the \NQLns{} LNS pairs. Every LM scores higher on the LLM-filtered QFrBLiMP than on the
human-filtered one (mean \QLmH{} points, \QLmHCI). On BLiMP, where the filter removes only \BRemovedLlm{} pairs, the two
filtered versions give the same mean accuracy (\BLmH, \BLmHCI).

\paragraph{Rankings barely move.} The order of the \NLm{} LMs is stable across no filter, human filter and LLM filter
(Kendall $\tau \ge \TauMin$; at most \FlipMax{} of 45 model pairs swap). The harm is to absolute scores and to which
failures stay visible. On a benchmark filtered this way, LMs look closer to mastery than the human judgments warrant,
and many of the pairs that would show the gap are gone.

\section{A frozen replication in Dutch}
\label{sec:replication}

After Phase~1 we found a third public minimal-pair dataset with per-judge data, BLiMP-NL
\citep{suijkerbuijk2025blimpnl}. We used it to ask whether the downstream result holds in a new language and with a
different kind of human judgment. Before any validator or LM saw the data, we locked a third protocol that freezes
everything else. It keeps the three LLMs and the LLM panel, with prompts translated sentence by sentence into Dutch
(Appendix~\ref{app:prompts}). It also keeps the primary filter (Gemini RATE, not re-selected), the \NLm{} LMs and
their scoring, the estimands and the Phase~1 criterion.

\paragraph{Data and support.} We use the \NNlItems{} pairs of the small BLiMP-NL release that have human ratings,
from \NNlPhen{} paradigms. Native speakers of Dutch rated single sentences on a 7-point scale. Each rater saw only one
sentence of a pair; the median is \NlRatGood{} ratings per acceptable and \NlRatBad{} per unacceptable sentence.
Support $s$ is the probability that a random rating of the acceptable sentence is higher than a random rating of the
unacceptable one (per-rater $z$-scores, ties counted as half). This measures how the two rating distributions are
ordered. It is not a share of forced choices, so it is not the same measure as in BLiMP and QFrBLiMP. It gives
\NNlLns{} LNS pairs.

\paragraph{The downstream effect replicates.} Both pre-registered conditions hold (Table~\ref{tab:downstream},
right). Human-supported pairs that the primary filter flags are \NlDeltaAbs{} points harder for LMs than those it
accepts ($\bar\Delta = \NlDelta$, \NlDeltaCI; \NlFlagSup{} flagged supported pairs), and $\Delta<0$ for all
\NlLmNeg{} LMs. Removing the flagged pairs raises accuracy by \NlLift{} points more than random removal
(\NlLiftCI). The lift is smaller than in Phase~1 because the filter flags fewer pairs. The other six validators range
from \NlOthMax{} to \NlOthMin{} points; \NlHolmSig{} of \NlNOth{} are significant after Holm correction, and the
exception is the LLM panel, which flags only 7 supported pairs. Of the \NlCombo{} combinations of LM and validator,
\NlComboNeg{} are negative. The gap holds with length-normalised log-probabilities (\NlLen, \NlLenCI) and when any
paradigm is left out (\NlLooMin{} to \NlLooMax). On strongly supported pairs it has a similar size, but its interval
reaches zero (\NlStrong, \NlStrongCI).

\paragraph{Detection and controls.} Detection looks as in Phase~0 (Table~\ref{tab:nl}). No validator matches more
than one rater per sentence. Forced-choice validators recall \NlFcRecMin--\NlFcRecMax{} of the LNS pairs, against
\NlPanelRecA{} for one rater. The best validator, \NlBest, reaches an F1 of \NlBestF{}, against \NlPanelFB{} for three
raters. Because a rater here judges one sentence, not a pair, we treat this comparison as descriptive. Two secondary
results are weaker than in Phase~1. First, pairs flagged by a single rater are also hard for LMs (\NlCtlHOne,
\NlCtlHOneCI). LLM-flagged pairs are harder still for all \NlCtlVsOneNeg{} validators by point estimate, but the
difference has an interval below zero for only \NlCtlVsOneSig. Second, LLM-filtered and human-filtered BLiMP-NL give
the same mean accuracy (\NlLmH, \NlLmHCI), as on BLiMP.

\section{Related work}
\label{sec:related}

\paragraph{Validating minimal pairs.} BLiMP validated a sample of its pairs with crowd raters
\citep{warstadt2020blimp}, and QFrBLiMP had native speakers judge every pair \citep{beauchemin2026qfrblimp}. RuBLiMP
used student annotators \citep{taktasheva2024rublimp}. CxMP reports a large human study \citep{oba2026cxmp}, and
FreqBLiMP reports a human audit of its stimuli \citep{white2026freqblimp}. MultiBLiMP builds
pairs for 101 languages from treebanks, without human validation \citep{jumelet2025multiblimp}. Work on acceptability
shows that judgments are graded and that informal linguist judgments mostly, but not always, replicate in formal
experiments \citep{sprouse2013comparison,lau2017grammaticality}. We take these points as given and ask what happens when
an LLM replaces the human step.

\paragraph{LLMs as annotators and judges.} LLMs can match or beat crowd workers on text annotation
\citep{gilardi2023chatgpt} and serve as judges of generated text \citep{zheng2023judging}. They also favour their own
outputs \citep{panickssery2024llm}. For fairness benchmarks, \citet{felkner2024gpt} find that LLM annotation cannot
replace human annotation. \citet{calderon2025alternative} propose the Alternative Annotator Test, which asks whether
an LLM aligns with a group of human annotators at least as well as a held-out member of the group does. Our equivalent panel size is in the same
spirit. We add two things: a target, the pairs whose label humans reject, rather than overall agreement; and a test of
what the substitution does to the conclusions drawn from the benchmark.

\paragraph{LLM grammaticality judgments.} \citet{dentella2023systematic} report low accuracy and a yes-bias when LLMs
judge grammaticality. \citet{hu2024language} re-test the same items with prompts like the human task and find that
LLMs match the intended labels better than humans, arguing that human ``errors'' often reflect real variation in
acceptability. Our results fit the second finding and show its cost. A validator that is more faithful to the
intended label than humans are will not report where humans disagree with it.

\section{Discussion}
\label{sec:discussion}

\paragraph{Why LLM validation may favour LMs.} One explanation, which our correlational data do not test directly,
is that an LLM validator and an LM under test learn from similar text. A sentence that an LLM rates as unnatural would
then tend to be one that LMs assign low probability. When the intended sentence of a pair is such a sentence, the LLM
flags the pair and the LM gets it wrong, even though humans accept it. Removing flagged pairs then selects the
benchmark on the outcome it measures. The mirror case adds to this. LNS pairs that the validator keeps stay in the
benchmark, and the LM is graded against a label that humans do not hold. Both effects push the benchmark toward the
models' preferences and away from the speakers'. Table~\ref{tab:examples} shows typical cases. In two QFrBLiMP pairs, almost all native judges choose
the intended word, while the validators and most LMs prefer the other word. In the other two, half of
the judges reject the intended label, plausibly because both variants are in use, and no validator flags the pair.
Rare words are one likely source of the shared preference \citep{white2026freqblimp}, but we do not separate the
sources here.

\paragraph{Recommendations.} For minimal pairs, an LLM should not replace human judges.
\begin{itemize}\setlength\itemsep{0pt}
\item \textbf{Do not drop LLM-flagged pairs unchecked.} If an LLM pre-screens a benchmark, send the pairs it flags
to human judges instead of deleting them. Many will be valid, and they are the hard ones.
\item \textbf{Do not read LLM agreement as validation.} High agreement with the intended label is what a validator that
rarely disagrees would show. Report recall on pairs that humans reject, which requires some human judgments.
\item \textbf{Use several human judges.} Single judges are noisy, but in our data a panel of three has a higher
F1 than every LLM validator on both datasets (by point estimate).
\item \textbf{Report how the labels were validated}, and release per-judge data so that others can repeat analyses
like ours.
\end{itemize}

\section*{Limitations}

\begin{itemize}\setlength\itemsep{0pt}
\item \textbf{Three datasets.} We found only three public minimal-pair datasets that release individual judgments. BLiMP's validated
subset has \NBLns{} LNS pairs, so the Phase~0 comparison on BLiMP has wide intervals and its pre-registered verdict
is a weak pass. QFrBLiMP has more LNS pairs, but its split-half reliability is low (\RelQ), so it was not used for
that verdict. The downstream effect is large on all three datasets. On BLiMP-NL, support comes from ratings of single
sentences, so its detection results are only descriptive, and its human-flag control is weaker than in Phase~1.
\item \textbf{LNS is not invalidity.} A pair humans reject may reflect dialect, graded acceptability or the task. We
claim only that LLM validators report few of the pairs humans reject, and that the pairs they do flag are hard for
LMs.
\item \textbf{Validators.} Three API models, three prompt formats (one added post hoc) and one temperature. Other prompts, chain-of-thought or
in-context examples may change detection. The models may change behind their APIs, so we release all outputs.
\item \textbf{LMs.} Ten open LMs up to 9B parameters, scored by total log-probability. DeepSeek-7B is near chance on
QFrBLiMP (\DsQNone) and has the smallest gap. We do not test larger or closed models.
\item \textbf{Mechanism.} The analysis is correlational. We show that LLM flags and LM errors coincide on
human-supported pairs, not why.
\end{itemize}

\section*{Ethics statement}

We use only public data under their licences: BLiMP (CC BY 4.0), QFrBLiMP (CC BY-NC-SA 4.0) and BLiMP-NL
(CC BY-SA 4.0). We collected no new
human judgments. We release code, LLM validator outputs and LM scores keyed by item, but not the per-judge
annotations, which remain available from the original authors. LLM outputs are never presented as human judgments.

\bibliography{refs}

\begin{thebibliography}{16}
\providecommand{\natexlab}[1]{#1}

\bibitem[{Beauchemin et~al.(2026)Beauchemin, Veilleux, Roy, and
  Khoury}]{beauchemin2026qfrblimp}
David Beauchemin, Pier-Luc Veilleux, Johanna-Pascale Roy, and Richard Khoury.
  2026.
\newblock \href {https://arxiv.org/abs/2509.25664} {{QFrBLiMP: a Quebec-French
  Benchmark of Linguistic Minimal Pairs}}.
\newblock \emph{Preprint}, arXiv:2509.25664.

\bibitem[{Calderon et~al.(2025)Calderon, Reichart, and
  Dror}]{calderon2025alternative}
Nitay Calderon, Roi Reichart, and Rotem Dror. 2025.
\newblock \href {https://doi.org/10.18653/v1/2025.acl-long.782} {The
  alternative annotator test for {LLM}-as-a-judge: How to statistically justify
  replacing human annotators with {LLM}s}.
\newblock In \emph{Proceedings of the 63rd Annual Meeting of the {A}ssociation
  for {C}omputational {L}inguistics (Volume 1: Long Papers)}, pages
  16051--16081, Vienna, Austria. Association for Computational Linguistics.

\bibitem[{Dentella et~al.(2023)Dentella, Günther, and
  Leivada}]{dentella2023systematic}
Vittoria Dentella, Fritz Günther, and Evelina Leivada. 2023.
\newblock \href {https://doi.org/10.1073/pnas.2309583120} {{Systematic testing
  of three Language Models reveals low language accuracy, absence of response
  stability, and a yes-response bias}}.
\newblock \emph{Proceedings of the National Academy of Sciences}, 120(51).

\bibitem[{Felkner et~al.(2024)Felkner, Thompson, and May}]{felkner2024gpt}
Virginia~K. Felkner, Jennifer~A. Thompson, and Jonathan May. 2024.
\newblock \href {https://arxiv.org/abs/2405.15760} {{GPT is Not an Annotator:
  The Necessity of Human Annotation in Fairness Benchmark Construction}}.
\newblock \emph{Preprint}, arXiv:2405.15760.

\bibitem[{Gilardi et~al.(2023)Gilardi, Alizadeh, and
  Kubli}]{gilardi2023chatgpt}
Fabrizio Gilardi, Meysam Alizadeh, and Maël Kubli. 2023.
\newblock \href {https://doi.org/10.1073/pnas.2305016120} {{ChatGPT outperforms
  crowd workers for text-annotation tasks}}.
\newblock \emph{Proceedings of the National Academy of Sciences}, 120(30).

\bibitem[{Hu et~al.(2024)Hu, Mahowald, Lupyan, Ivanova, and
  Levy}]{hu2024language}
Jennifer Hu, Kyle Mahowald, Gary Lupyan, Anna Ivanova, and Roger Levy. 2024.
\newblock \href {https://doi.org/10.1073/pnas.2400917121} {{Language models
  align with human judgments on key grammatical constructions}}.
\newblock \emph{Proceedings of the National Academy of Sciences}, 121(36).

\bibitem[{Jumelet et~al.(2026)Jumelet, Weissweiler, Nivre, and
  Bisazza}]{jumelet2025multiblimp}
Jaap Jumelet, Leonie Weissweiler, Joakim Nivre, and Arianna Bisazza. 2026.
\newblock \href {https://arxiv.org/abs/2504.02768} {{MultiBLiMP 1.0: A
  Massively Multilingual Benchmark of Linguistic Minimal Pairs}}.
\newblock \emph{Preprint}, arXiv:2504.02768.

\bibitem[{Lau et~al.(2017)Lau, Clark, and Lappin}]{lau2017grammaticality}
Jey~Han Lau, Alexander Clark, and Shalom Lappin. 2017.
\newblock \href {https://doi.org/10.1111/cogs.12414} {{Grammaticality,
  Acceptability, and Probability: A Probabilistic View of Linguistic
  Knowledge}}.
\newblock \emph{Cognitive Science}, 41(5):1202--1241.

\bibitem[{Oba and Sugawara(2026)}]{oba2026cxmp}
Miyu Oba and Saku Sugawara. 2026.
\newblock \href {https://arxiv.org/abs/2602.21978} {{CxMP: A Linguistic
  Minimal-Pair Benchmark for Evaluating Constructional Understanding in
  Language Models}}.
\newblock \emph{Preprint}, arXiv:2602.21978.

\bibitem[{Panickssery et~al.(2024)Panickssery, Bowman, and
  Feng}]{panickssery2024llm}
Arjun Panickssery, Samuel~R. Bowman, and Shi Feng. 2024.
\newblock \href {https://arxiv.org/abs/2404.13076} {{LLM Evaluators Recognize
  and Favor Their Own Generations}}.
\newblock \emph{Preprint}, arXiv:2404.13076.

\bibitem[{Sprouse et~al.(2013)Sprouse, Schütze, and
  Almeida}]{sprouse2013comparison}
Jon Sprouse, Carson~T. Schütze, and Diogo Almeida. 2013.
\newblock \href {https://doi.org/10.1016/j.lingua.2013.07.002} {{A comparison
  of informal and formal acceptability judgments using a random sample from
  Linguistic Inquiry 2001--2010}}.
\newblock \emph{Lingua}, 134:219--248.

\bibitem[{Suijkerbuijk et~al.(2025)Suijkerbuijk, Prins, Kloots, Zuidema, and
  Frank}]{suijkerbuijk2025blimpnl}
Michelle Suijkerbuijk, Zo{\"e} Prins, Marianne de~Heer Kloots, Willem Zuidema,
  and Stefan~L. Frank. 2025.
\newblock \href {https://doi.org/10.1162/coli_a_00559} {{BLiMP-NL}: A corpus of
  {D}utch minimal pairs and acceptability judgments for language model
  evaluation}.
\newblock \emph{Computational Linguistics}, 51(4):1267--1301.

\bibitem[{Taktasheva et~al.(2024)Taktasheva, Bazhukov, Koncha, Fenogenova,
  Artemova, and Mikhailov}]{taktasheva2024rublimp}
Ekaterina Taktasheva, Maxim Bazhukov, Kirill Koncha, Alena Fenogenova,
  Ekaterina Artemova, and Vladislav Mikhailov. 2024.
\newblock \href {https://doi.org/10.18653/v1/2024.emnlp-main.522}
  {{R}u{BL}i{MP}: {R}ussian benchmark of linguistic minimal pairs}.
\newblock In \emph{Proceedings of the 2024 Conference on Empirical Methods in
  Natural Language Processing}, pages 9268--9299, Miami, Florida, USA.
  Association for Computational Linguistics.

\bibitem[{Warstadt et~al.(2020)Warstadt, Parrish, Liu, Mohananey, Peng, Wang,
  and Bowman}]{warstadt2020blimp}
Alex Warstadt, Alicia Parrish, Haokun Liu, Anhad Mohananey, Wei Peng, Sheng-Fu
  Wang, and Samuel~R. Bowman. 2020.
\newblock \href {https://doi.org/10.1162/tacl_a_00321} {{BL}i{MP}: The
  benchmark of linguistic minimal pairs for {E}nglish}.
\newblock \emph{Transactions of the Association for Computational Linguistics},
  8:377--392.

\bibitem[{White and Arase(2026)}]{white2026freqblimp}
Tyrone White and Yuki Arase. 2026.
\newblock \href {https://arxiv.org/abs/2609.07153} {{FreqBLiMP:
  Frequency-Controlled Minimal Pairs Reveal Robustness and Fragility of LLMs
  Under Lexical Rarity}}.
\newblock \emph{Preprint}, arXiv:2609.07153.

\bibitem[{Zheng et~al.(2023)Zheng, Chiang, Sheng, Zhuang, Wu, Zhuang, Lin, Li,
  Li, Xing, Zhang, Gonzalez, and Stoica}]{zheng2023judging}
Lianmin Zheng, Wei-Lin Chiang, Ying Sheng, Siyuan Zhuang, Zhanghao Wu, Yonghao
  Zhuang, Zi~Lin, Zhuohan Li, Dacheng Li, Eric~P. Xing, Hao Zhang, Joseph~E.
  Gonzalez, and Ion Stoica. 2023.
\newblock \href {https://arxiv.org/abs/2306.05685} {{Judging LLM-as-a-Judge
  with MT-Bench and Chatbot Arena}}.
\newblock \emph{Preprint}, arXiv:2306.05685.

\end{thebibliography}

\appendix
\section{Validator prompts}
\label{app:prompts}

System prompts, verbatim. The user message lists 12 numbered pairs (FC) or 12 numbered sentences (RATE). An answer
is accepted if it contains a JSON array of the right length and range; otherwise the request is retried up to five
times.

\paragraph{FC, English.} \emph{You are a native speaker of English. Read pairs of English sentences and choose the
more natural one. Go with your first impression. Output only a JSON array with one entry per pair, 1 or 2 for the
sentence you choose, in the order of the pair numbers. No explanation.}

\paragraph{FC, French.} \emph{Vous êtes locuteur natif du français québécois. Pour chaque paire, sélectionnez la
phrase que vous jugez bien écrite parmi les deux phrases disponibles. Fiez-vous à votre première impression.
Répondez uniquement par un tableau JSON avec une entrée par paire, 1 ou 2 selon la phrase choisie, dans l'ordre des
numéros de paire. Aucune explication.}

\paragraph{RATE, English.} \emph{You are a native speaker of English rating sentences by feel. Go with your first
impression. How acceptable is each sentence as natural English? 1 = completely impossible, 2 = very awkward,
3 = marginal, 4 = mostly natural, 5 = completely natural. Rate each sentence on its own. Output only a JSON array of
integers in the order of the sentence numbers, no explanation.}

\paragraph{RATE, French.} \emph{Vous êtes locuteur natif du français québécois et vous évaluez des phrases à
l'intuition. Fiez-vous à votre première impression. Dans quelle mesure chaque phrase est-elle acceptable en
français ? 1 = complètement impossible, 2 = très maladroite, 3 = marginale, 4 = plutôt naturelle, 5 = tout à fait
naturelle. Évaluez chaque phrase séparément. Répondez uniquement par un tableau JSON d'entiers dans l'ordre des
numéros, sans explication.}

\paragraph{Dutch (Section~\ref{sec:replication}).} Translated sentence by sentence from the English prompts, locked
before any output. FC: \emph{U bent moedertaalspreker van het Nederlands. Lees paren Nederlandse zinnen en kies
telkens de meest natuurlijke zin. Ga uit van uw eerste indruk. Geef alleen een JSON-array met één waarde per paar, 1
of 2 voor de zin die u kiest, in de volgorde van de paarnummers. Geen uitleg.} RATE: \emph{U bent moedertaalspreker
van het Nederlands en beoordeelt zinnen op gevoel. Ga uit van uw eerste indruk. Hoe acceptabel is elke zin als
natuurlijk Nederlands? 1 = volkomen onmogelijk, 2 = erg onbeholpen, 3 = twijfelachtig, 4 = grotendeels natuurlijk,
5 = volkomen natuurlijk. Beoordeel elke zin op zichzelf. Geef alleen een JSON-array met gehele getallen in de
volgorde van de zinsnummers, zonder uitleg.} VERIFY: \emph{U bent moedertaalspreker van het Nederlands en controleert
de labels van een taalkundige benchmark. In elk paar is zin A gelabeld als acceptabel en zin B als onacceptabel.
Zouden de meeste moedertaalsprekers van het Nederlands het ermee eens zijn dat A acceptabel is en natuurlijker dan B?
Antwoord 1 als ze het met het label eens zouden zijn, 0 als dat niet zo is. Geef alleen een JSON-array met één
waarde per paar, 1 of 0, in de volgorde van de paarnummers. Geen uitleg.}

\section{VERIFY prompt}
\label{app:verify}

Added after Phase~1 and logged, with its analysis, before any output existed. Exploratory: it does not replace the
primary filter or change any verdict. Each request lists 12 pairs as ``A (acceptable): \ldots{} B (unacceptable):
\ldots''. A pair is flagged if the answer is 0. Every request returned a parseable answer. Table~\ref{tab:verify}
gives the results.

\paragraph{VERIFY, English.} \emph{You are a native speaker of English checking the labels of a linguistics benchmark.
In each pair, sentence A is labelled acceptable and sentence B is labelled unacceptable. Would most native speakers of
English agree that A is acceptable and more natural than B? Answer 1 if they would agree with the label, 0 if not.
Output only a JSON array with one entry per pair, 1 or 0, in the order of the pair numbers. No explanation.}

\paragraph{VERIFY, French.} \emph{Vous êtes locuteur natif du français québécois et vous vérifiez les étiquettes d'un
banc d'essai linguistique. Dans chaque paire, la phrase A est étiquetée acceptable et la phrase B inacceptable. La
plupart des locuteurs natifs du français québécois seraient-ils d'accord que A est acceptable et plus naturelle que
B ? Répondez 1 s'ils seraient d'accord avec l'étiquette, 0 sinon. Répondez uniquement par un tableau JSON avec une
entrée par paire, 1 ou 0, dans l'ordre des numéros de paire. Aucune explication.}

\begin{table*}[t]
\centering
\small
\resizebox{\textwidth}{!}{%
\begin{tabular}{l ccccc ccccc cc}
\toprule
 & \multicolumn{5}{c}{BLiMP} & \multicolumn{5}{c}{QFrBLiMP} & \multicolumn{2}{c}{Open LMs} \\
\cmidrule(lr){2-6}\cmidrule(lr){7-11}\cmidrule(lr){12-13}
Validator & Rec & Prec & F1 & Agree & $k_{\text{eq}}$ & Rec & Prec & F1 & Agree & $k_{\text{eq}}$ & $\bar\Delta$ [95\% CI] & Lift$-$rand. \\
\midrule
1 judge & 0.47 & 0.16 & 0.239 & 0.869 & -- & 0.50 & 0.22 & 0.304 & 0.862 & -- & -- & -- \\
3 judges & 0.46 & 0.26 & 0.331 & 0.919 & -- & 0.50 & 0.35 & 0.409 & 0.913 & -- & -- & -- \\
5 judges & 0.44 & 0.32 & 0.365 & 0.934 & -- & 0.49 & 0.43 & 0.454 & 0.929 & -- & -- & -- \\
\midrule
Gemini 3.8 Flash VERIFY & 0.40 & 0.18 & 0.244 & 0.891 & 1 & 0.67 & 0.19 & 0.299 & 0.809 & $<$1 & \ensuremath{-}17.8 [\ensuremath{-}21.1, \ensuremath{-}14.6] & +4.4 \\
Claude Sonnet 4.6 VERIFY & 0.28 & 0.19 & 0.224 & 0.917 & $<$1 & 0.47 & 0.24 & 0.321 & 0.881 & 1 & \ensuremath{-}18.9 [\ensuremath{-}24.0, \ensuremath{-}13.9] & +2.7 \\
GPT-6 Luna VERIFY & 0.33 & 0.10 & 0.153 & 0.841 & $<$1 & 0.62 & 0.14 & 0.229 & 0.748 & $<$1 & \ensuremath{-}14.2 [\ensuremath{-}17.0, \ensuremath{-}11.4] & +4.5 \\
\bottomrule
\end{tabular}}
\caption{VERIFY prompt (post hoc, exploratory): the intended label is shown and the LLM answers whether most native speakers would agree. Columns as in Tables~\ref{tab:phase0} and~\ref{tab:downstream}; human panels are the same draws as in Table~\ref{tab:phase0}. Lift$-$rand.: score lift from removing flagged pairs minus lift from removing as many random pairs, in points.}
\label{tab:verify}
\end{table*}

\section{Secondary and robustness analyses}
\label{app:secondary}

\begin{table}[t]
\centering
\small
\resizebox{\columnwidth}{!}{%
\begin{tabular}{l ccc ccc}
\toprule
 & \multicolumn{3}{c}{BLiMP (50)} & \multicolumn{3}{c}{QFrBLiMP (255)} \\
\cmidrule(lr){2-4}\cmidrule(lr){5-7}
Validator & Rec & F1 & F1 $-$ 3 judges & Rec & F1 & F1 $-$ 3 judges \\
\midrule
1 judge & 0.34 & 0.405 & -- & 0.31 & 0.391 & -- \\
3 judges & 0.68 & 0.525 & -- & 0.60 & 0.529 & -- \\
5 judges & 0.54 & 0.579 & -- & 0.44 & 0.520 & -- \\
\midrule
Gemini 3.8 Flash FC & 0.05 & 0.095 & [\ensuremath{-}0.56, \ensuremath{-}0.29] & 0.25 & 0.376 & [\ensuremath{-}0.21, \ensuremath{-}0.10] \\
Gemini 3.8 Flash RATE & 0.27 & 0.364 & [\ensuremath{-}0.30, \ensuremath{-}0.02] & 0.54 & 0.524 & [\ensuremath{-}0.05, +0.04] \\
Claude Sonnet 4.6 FC & 0.13 & 0.215 & [\ensuremath{-}0.45, \ensuremath{-}0.16] & 0.26 & 0.373 & [\ensuremath{-}0.21, \ensuremath{-}0.09] \\
Claude Sonnet 4.6 RATE & 0.46 & 0.472 & [\ensuremath{-}0.18, +0.08] & 0.50 & 0.513 & [\ensuremath{-}0.06, +0.03] \\
GPT-6 Luna FC & 0.14 & 0.236 & [\ensuremath{-}0.44, \ensuremath{-}0.14] & 0.24 & 0.351 & [\ensuremath{-}0.23, \ensuremath{-}0.12] \\
GPT-6 Luna RATE & 0.43 & 0.435 & [\ensuremath{-}0.21, +0.03] & 0.54 & 0.490 & [\ensuremath{-}0.08, +0.01] \\
LLM panel (3$\times$FC) & 0.08 & 0.149 & [\ensuremath{-}0.51, \ensuremath{-}0.23] & 0.22 & 0.336 & [\ensuremath{-}0.25, \ensuremath{-}0.14] \\
\bottomrule
\end{tabular}}
\caption{Secondary gold: weak support ($s'<0.75$). Numbers of weakly supported pairs (full judge set) in brackets. Intervals: 95\% paired item bootstrap.}
\label{tab:weak}
\end{table}

\paragraph{Weak-support gold.} Table~\ref{tab:weak} repeats the Phase~0 comparison with $s'<0.75$ as the gold. Every
forced-choice validator and the LLM panel fall below the three-judge panel, with intervals that exclude zero on both
datasets. Rating validators come closer.

\paragraph{Strict rating flag.} Flagging a pair only when the intended sentence is rated strictly lower
($r_{\text{good}} < r_{\text{bad}}$) lowers the recall of rating validators and leaves the conclusions unchanged
(supplement).

\paragraph{Human-flag control.} Table~\ref{tab:control} gives the full results of the control in
Section~\ref{sec:downstream}. It was added after Phase~1 and logged before it was run, and it is exploratory.

\begin{table*}[t]
\centering
\small
\begin{tabular}{l r r c c c}
\toprule
Flag source & Flagged (\%) & $\Delta$ & 95\% CI & minus 1 judge & minus 3 judges \\
\midrule
1 judge & 11.4 & \ensuremath{-}13.7 & [\ensuremath{-}18.1, \ensuremath{-}10.0] & -- & -- \\
3 judges & 5.9 & \ensuremath{-}21.7 & [\ensuremath{-}27.6, \ensuremath{-}16.4] & -- & -- \\
\midrule
Gemini 3.8 Flash FC & 4.6 & \ensuremath{-}30.4 & [\ensuremath{-}36.7, \ensuremath{-}23.7] & \ensuremath{-}16.7 [\ensuremath{-}23.7, \ensuremath{-}9.6] & \ensuremath{-}8.7 [\ensuremath{-}16.2, \ensuremath{-}1.1] \\
Gemini 3.8 Flash RATE & 19.4 & \ensuremath{-}19.5 & [\ensuremath{-}22.6, \ensuremath{-}16.2] & \ensuremath{-}5.8 [\ensuremath{-}10.1, \ensuremath{-}1.3] & +2.2 [\ensuremath{-}3.4, +8.4] \\
Claude Sonnet 4.6 FC & 6.2 & \ensuremath{-}29.7 & [\ensuremath{-}35.6, \ensuremath{-}24.2] & \ensuremath{-}16.0 [\ensuremath{-}22.1, \ensuremath{-}9.8] & \ensuremath{-}8.0 [\ensuremath{-}14.9, \ensuremath{-}0.5] \\
Claude Sonnet 4.6 RATE & 19.6 & \ensuremath{-}22.2 & [\ensuremath{-}25.5, \ensuremath{-}19.2] & \ensuremath{-}8.4 [\ensuremath{-}12.8, \ensuremath{-}3.7] & \ensuremath{-}0.5 [\ensuremath{-}6.1, +5.7] \\
GPT-6 Luna FC & 5.7 & \ensuremath{-}27.5 & [\ensuremath{-}33.5, \ensuremath{-}21.5] & \ensuremath{-}13.7 [\ensuremath{-}20.2, \ensuremath{-}7.2] & \ensuremath{-}5.8 [\ensuremath{-}12.6, +1.8] \\
GPT-6 Luna RATE & 24.6 & \ensuremath{-}18.9 & [\ensuremath{-}21.9, \ensuremath{-}16.0] & \ensuremath{-}5.2 [\ensuremath{-}9.4, \ensuremath{-}0.7] & +2.8 [\ensuremath{-}2.5, +9.1] \\
LLM panel (3$\times$FC) & 4.3 & \ensuremath{-}32.2 & [\ensuremath{-}39.6, \ensuremath{-}24.9] & \ensuremath{-}18.5 [\ensuremath{-}25.9, \ensuremath{-}11.2] & \ensuremath{-}10.5 [\ensuremath{-}18.5, \ensuremath{-}2.8] \\
\bottomrule
\end{tabular}
\caption{Human-flag control (post hoc, exploratory). In each judge reshuffle, the first judge or the majority of the first three flags a pair, and the held-out support $s'$ comes from judges 6 onward. On pairs with $s'>0.5$, $\Delta$ is LM accuracy on flagged minus unflagged pairs, in points, averaged over LMs and combined over datasets with the primary weights. Flagged: share of these pairs flagged. Last two columns: validator $\Delta$ minus human $\Delta$, with 95\% item-bootstrap CIs.}
\label{tab:control}
\end{table*}

\paragraph{BLiMP-NL.} Table~\ref{tab:nl} gives the per-validator results of the replication in
Section~\ref{sec:replication}.

\begin{table*}[t]
\centering
\small
\resizebox{\textwidth}{!}{%
\begin{tabular}{l cc c c c c c}
\toprule
 & \multicolumn{3}{c}{LNS detection (descriptive)} & \multicolumn{2}{c}{Downstream} & \multicolumn{2}{c}{Human-flag control} \\
\cmidrule(lr){2-4}\cmidrule(lr){5-6}\cmidrule(lr){7-8}
Validator & Rec & F1 & $k_{\text{eq}}$ & Flagged (sup / all) & $\bar\Delta$ [95\% CI] & $\Delta$ & $-$ 1 rater [95\% CI] \\
\midrule
1 rater per sentence & 0.49 & 0.294 & -- & -- & -- & \ensuremath{-}8.5 [\ensuremath{-}14.7, \ensuremath{-}2.7] & -- \\
3 raters per sentence & 0.50 & 0.430 & -- & -- & -- & \ensuremath{-}15.6 [\ensuremath{-}25.5, \ensuremath{-}5.5] & -- \\
5 raters per sentence & 0.51 & 0.480 & -- & -- & -- & -- & -- \\
\midrule
Gemini 3.8 Flash FC & 0.15 & 0.237 & $<$1 & 6 / 13 & \ensuremath{-}47.0 [\ensuremath{-}64.8, \ensuremath{-}25.0] & \ensuremath{-}47.5 & \ensuremath{-}39.0 [\ensuremath{-}58.5, \ensuremath{-}16.6] \\
Gemini 3.8 Flash RATE$^\dagger$ & 0.49 & 0.388 & 1 & 50 / 69 & \ensuremath{-}13.4 [\ensuremath{-}21.1, \ensuremath{-}5.4] & \ensuremath{-}13.5 & \ensuremath{-}5.0 [\ensuremath{-}14.5, +4.2] \\
Claude Sonnet 4.6 FC & 0.43 & 0.372 & 1 & 41 / 60 & \ensuremath{-}16.7 [\ensuremath{-}25.4, \ensuremath{-}8.0] & \ensuremath{-}17.0 & \ensuremath{-}8.5 [\ensuremath{-}19.4, +0.9] \\
Claude Sonnet 4.6 RATE & 0.62 & 0.304 & 1 & 114 / 140 & \ensuremath{-}11.4 [\ensuremath{-}17.1, \ensuremath{-}6.0] & \ensuremath{-}11.4 & \ensuremath{-}2.9 [\ensuremath{-}10.2, +3.8] \\
GPT-6 Luna FC & 0.17 & 0.195 & $<$1 & 29 / 36 & \ensuremath{-}11.0 [\ensuremath{-}20.1, \ensuremath{-}2.1] & \ensuremath{-}9.7 & \ensuremath{-}1.2 [\ensuremath{-}12.2, +9.4] \\
GPT-6 Luna RATE & 0.53 & 0.282 & $<$1 & 102 / 126 & \ensuremath{-}8.7 [\ensuremath{-}14.3, \ensuremath{-}3.1] & \ensuremath{-}8.9 & \ensuremath{-}0.4 [\ensuremath{-}7.9, +6.8] \\
LLM panel (3$\times$FC) & 0.19 & 0.279 & $<$1 & 7 / 16 & \ensuremath{-}19.2 [\ensuremath{-}40.9, +1.4] & \ensuremath{-}22.9 & \ensuremath{-}14.4 [\ensuremath{-}38.1, +9.2] \\
\midrule
Gemini 3.8 Flash VERIFY$^\ddagger$ & 0.44 & 0.345 & 1 & 51 / 70 & \ensuremath{-}21.0 [\ensuremath{-}30.2, \ensuremath{-}12.4] & \ensuremath{-}21.5 & \ensuremath{-}13.0 [\ensuremath{-}22.7, \ensuremath{-}2.9] \\
Claude Sonnet 4.6 VERIFY$^\ddagger$ & 0.44 & 0.303 & 1 & 68 / 87 & \ensuremath{-}2.1 [\ensuremath{-}8.6, +4.1] & \ensuremath{-}2.6 & +5.8 [\ensuremath{-}2.1, +13.6] \\
GPT-6 Luna VERIFY$^\ddagger$ & 0.56 & 0.272 & $<$1 & 118 / 140 & \ensuremath{-}10.1 [\ensuremath{-}15.4, \ensuremath{-}4.8] & \ensuremath{-}10.6 & \ensuremath{-}2.2 [\ensuremath{-}9.2, +5.5] \\
\bottomrule
\end{tabular}}
\caption{Frozen replication on BLiMP-NL (756 rated pairs, 38 LNS by rating-rank support). LNS detection follows Phase~0, with human panels of $k$ raters per sentence; since each rater saw one sentence of a pair, this is a descriptive comparison across measurement formats and $k_{\text{eq}}$ is a numeric reference only. Downstream: as in Table~\ref{tab:downstream}, in points over the same LMs. Human-flag control: $\Delta$ for pairs flagged by the validator or by one rater per sentence, with support from the held-out raters, and the difference between them. $^\dagger$Primary filter. $^\ddagger$VERIFY prompt (post hoc in Phase 0, locked here before any output).}
\label{tab:nl}
\end{table*}

\paragraph{Per-LM results.} Table~\ref{tab:lms} lists $\Delta$ and accuracy under each filter for every LM.

\begin{table*}[t]
\centering
\small
\begin{tabular}{l r c ccc ccc}
\toprule
 & & & \multicolumn{3}{c}{BLiMP accuracy} & \multicolumn{3}{c}{QFrBLiMP accuracy} \\
\cmidrule(lr){4-6}\cmidrule(lr){7-9}
LM & $\Delta$ & 95\% CI & none & human & LLM & none & human & LLM \\
\midrule
Qwen2.5-1.5B-Instruct & \ensuremath{-}21.5 & [\ensuremath{-}26.5, \ensuremath{-}16.6] & 0.802 & 0.827 & 0.830 & 0.852 & 0.882 & 0.920 \\
Qwen2.5-3B-Instruct & \ensuremath{-}22.6 & [\ensuremath{-}27.8, \ensuremath{-}17.5] & 0.779 & 0.805 & 0.804 & 0.801 & 0.826 & 0.867 \\
Qwen2.5-7B-Instruct & \ensuremath{-}20.5 & [\ensuremath{-}25.1, \ensuremath{-}15.9] & 0.784 & 0.805 & 0.820 & 0.879 & 0.910 & 0.944 \\
Llama-3.2-3B-Instruct & \ensuremath{-}20.6 & [\ensuremath{-}25.6, \ensuremath{-}15.7] & 0.770 & 0.805 & 0.804 & 0.840 & 0.865 & 0.901 \\
Llama-3.1-8B-Instruct & \ensuremath{-}17.0 & [\ensuremath{-}21.6, \ensuremath{-}12.6] & 0.825 & 0.848 & 0.839 & 0.876 & 0.901 & 0.933 \\
GLM-4-9B-chat & \ensuremath{-}23.4 & [\ensuremath{-}28.3, \ensuremath{-}18.5] & 0.799 & 0.821 & 0.823 & 0.858 & 0.887 & 0.928 \\
Mistral-7B & \ensuremath{-}20.0 & [\ensuremath{-}24.7, \ensuremath{-}15.2] & 0.825 & 0.845 & 0.839 & 0.868 & 0.894 & 0.931 \\
Yi-1.5-9B-Chat & \ensuremath{-}18.0 & [\ensuremath{-}22.8, \ensuremath{-}13.3] & 0.822 & 0.839 & 0.845 & 0.848 & 0.875 & 0.906 \\
DeepSeek-7B & \ensuremath{-}7.1 & [\ensuremath{-}12.8, \ensuremath{-}1.5] & 0.661 & 0.666 & 0.656 & 0.491 & 0.495 & 0.511 \\
Gemma-2-9B & \ensuremath{-}18.3 & [\ensuremath{-}23.0, \ensuremath{-}13.8] & 0.741 & 0.763 & 0.763 & 0.875 & 0.901 & 0.935 \\
\bottomrule
\end{tabular}
\caption{Per-LM results for the primary filter. $\Delta$ in points. Accuracy with no filter, after removing pairs natives do not support (human), and after removing pairs the primary LLM filter flags (LLM).}
\label{tab:lms}
\end{table*}

\section{Protocol notes}
\label{app:protocol}

All three protocols, with time stamps, are in the supplement. Deviations and clarifications:
\begin{itemize}\setlength\itemsep{0pt}
\item Phase~0, before any output was read: the resampling was clarified as nested panels (judges 1, 1--3, 1--5) and a
held-out gold from judges 6 onward. Reliability below 0.5 removes a dataset from the verdict, and pairs missing any
validator output are dropped (none were).
\item One RATE batch was split in halves after repeated empty responses (Section~\ref{sec:setup}). One validator run
was restarted after a slow request stalled its writer; it resumed without duplicates.
\item Two analyses were added after Phase~1, each logged with its method before it was run: the human-flag control
(Appendix~\ref{app:secondary}) and the VERIFY prompt (Appendix~\ref{app:verify}). Both are exploratory and change no
pre-registered verdict.
\item The BLiMP-NL protocol was locked after both post-hoc analyses and before any BLiMP-NL output. It includes the
VERIFY prompt and the human-flag control as secondary analyses. Had the primary filter flagged fewer than 10
supported pairs, the estimate would have been labelled low-information; it flagged \NlFlagSup.
\item Phase~1 adds one secondary interval that the protocol lists without specifying its method: the difference
between LLM-filtered and human-filtered accuracy, with the same item bootstrap.
\end{itemize}

\end{document}